\def\year{2021}\relax
\pdfoutput=1
\documentclass[letterpaper]{article} 
\usepackage{aaai21}  
\usepackage{times}  
\usepackage{helvet} 
\usepackage{courier}  
\usepackage[hyphens]{url}  
\usepackage{graphicx} 
\usepackage{natbib}  
\usepackage{caption} 
\usepackage[switch]{lineno}
\usepackage{comment}
\usepackage{color}
\usepackage{amsmath,amssymb} 
\usepackage{comment}
\usepackage{epsfig}
\usepackage{multirow}
\usepackage{makecell}
\usepackage{enumitem}

\title{Exploring the Hierarchy in Relation Labels for Scene Graph Generation}
\author {Yi Zhou\textsuperscript{\rm1, \rm 2}, Shuyang Sun\textsuperscript{\rm 3}, Chao Zhang\textsuperscript{\rm 2}, Yikang Li\textsuperscript{\rm 4}, Wanli Ouyang\textsuperscript{\rm 1} \\}
\affiliations {\textsuperscript{\rm 1}The University of Sydney, \textsuperscript{\rm 2}Samsung Research, \textsuperscript{\rm 3}University of Oxford, \textsuperscript{\rm 4}SenseTime Research \\ {\tt \small yi0813.zhou@samsung.com}}
\begin{document}

\maketitle
\begin{abstract}
By assigning each relationship a single label, current approaches formulate the relationship detection as a classification problem. Under this formulation, predicate categories are treated as completely different classes. However, different from the object labels where different classes have explicit boundaries, predicates usually have overlaps in their semantic meanings. 
For example, \emph{sit\_on} and \emph{stand\_on} have common meanings in vertical relationships but different details of how these two objects are vertically placed.
In order to leverage the inherent structures of the predicate categories, we propose to first build the language hierarchy and then
utilize the Hierarchy Guided Feature Learning (HGFL) strategy to learn better region features of both the coarse-grained level and the fine-grained level. Besides, we also propose the Hierarchy Guided Module (HGM) to utilize the 
coarse-grained level to guide the learning of 
fine-grained level features. 
Experiments show that the proposed simple yet effective method can improve several state-of-the-art baselines by a large margin (up to $33\%$ relative gain) 
in terms of Recall@50 on the task of Scene Graph Generation in different datasets.
\end{abstract}

\section{Introduction}
\label{sec:intro}
As a basic visual scene understanding task, scene graph generation
\cite{lu2016visual,xu2017scene,li2017vip,zhang2017visual,chen2018iterative,zellers2018neuralmotif,li2018factorizable,yang2018graph, chen2019knowledge} generates the scene graph including the located objects as nodes and the corresponding relationships between objects as edges from the image. 
Generally, the common solution for scene graph generation task is detecting the objects first and further inferring their pair-wise relationships, denoted as \emph{predicates}.
In this way, both the object detection and the predicate recognition are formulated as the
classification problems, where there is only one single ground truth label assigned to each instance,
assuming every category is
independent and orthogonal to each other. 

However, different from object labels, which can be clearly defined, 
the predicates' boundaries are fuzzy. Sometimes they have overlaps in semantics.
For example, "sitting next to" and "standing next to" are two different predicate labels but express similar spatial relations in semantics. 
Therefore, treating these semantic-overlapping predicates as independent classes fails to leverage these kinds of inherent structure of the predicate labels.

\begin{figure}
\centering
\includegraphics[scale=0.25]{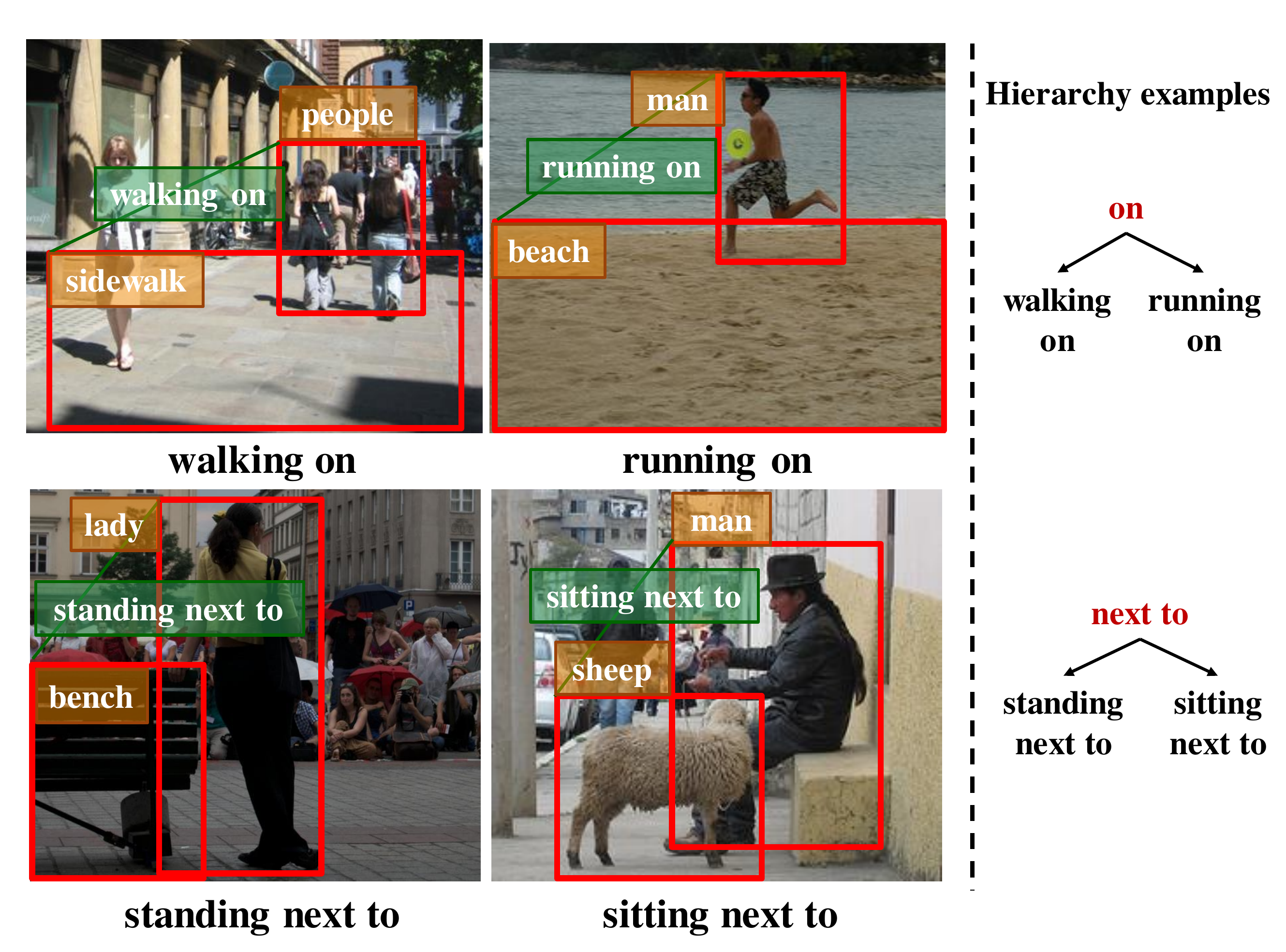}
\caption{\textbf{Examples of semantic-overlapping predicate labels
}.
\textbf{Left}: scene image with predicate labels. \textbf{Right}: constructed hierarchy examples.
Texts in red and bold are the clustered parent classes, which expresses the similarity connection between predicate labels. Best viewed in color.}
\label{fig:front_image}
\end{figure}

To explore the inherent connections in the predicate labels, 
a simple and direct solution is to cluster the predicate labels into parent classes to represent these semantic-overlapping connections. 
However, the semantic-overlapping connection is not the only property of predicate labels. Still take the "sitting next to" and "standing next to" as examples, as shown in the Figure \ref{fig:front_image}, although they are semantic-overlapping, they correspond to slightly different relative spatial positions in the image level. 
Therefore, the properties of both similarity and difference exist in the predicate labels. \emph{How to take both the similarity and difference into consideration?}

To solve the problem, we propose to 
build the hierarchy based on comprehensive semantic understanding
in two different perspectives, including human understanding and machine understanding.
The built hierarchy contains both the clustered parent classes and the cleaned predicate labels.
The cleaned predicate labels in the lower hierarchy serve as the fine-grained labels with richer descriptions, and the clustered parent classes in the higher hierarchy serve as the coarse-grained labels which contain the information of semantic-overlapping connections.

Based on the built hierarchy,
we propose the Hierarchy Guided Feature Learning (HGFL) strategy that learns better region features by simultaneously training two branches supervised by 
coarse-grained and fine-grained labels respectively.
Besides, we also propose the Hierarchy Guided Module (HGM) to better leverage the correlations between the coarse-grained and fine-grained region features.

Extensive experiments have shown that our method could remarkably improve the performance in terms of Recall@50 for about $18.9\%$ gain on Scene Graph Generation task by simply applying our HGFL strategy.
This strategy only requires marginal computational cost during training and no extra computational cost for inference.
Another at least $5\%$ gain in terms of Recall@50 on the Scene Graph Generation task could be achieved by embedding our HGM into the network.
To showcase that the proposed method is general and robust, experiments are performed on different datasets and different frameworks. 
Experiment results show that our method could outperform the state-of-the-art methods on the VG-MSDN \cite{li2017MSDN} dataset and the VG-DR-Net \cite{dai2017vg-drnet} dataset.
Besides, we validate that our method is general that could remarkably improve $4$ state-of-the-art frameworks.

The contributions of this paper are as follows:

(1) To better explore the inherent semantic-overlapping connections in predicate labels, we propose to build the predicate label hierarchy based on the \textbf{semantic meaning} of labels.

(2) To learn better region features on both coarse-grained and fine-grained levels, we introduce the Hierarchy Guided Feature Learning (HGFL) strategy.

(3) To better utilize the correlations between region features of coarse-grained and fine-grained levels, we further propose the Hierarchy Guided Module (HGM) to reason both the region-wise and channel-wise correlations and finally refine the region features with these two correlations.

\section{Related Work}
\textbf{Scene Graph Generation.}  
To improve the performance of scene graph generation task, recent works mainly focus on several perspectives such as solving the labeling problems in the dataset \cite{zhang2019graphical,zhan2019exploring,chen2019limitedlabels,peyre2019detectingunseen}, importing external text knowledge as conditions \cite{yu2017visualdistill,gu2019external}, exploring more information (motif, correlations) in existing labels \cite{zellers2018neuralmotif,chen2019counterfactual,wang2019exploring_context} and increasing the diversity of feature information (\emph{e.g.} visual feature, spatial features, linguistic features etc.) \cite{qi2019attentive,dai2017vg-drnet,hung2019union}.

Specifically, there are several labeling problems such as ambiguous labels or instances, imbalanced classes and incomplete annotations existing in the Visual Genome Dataset \cite{krishna2017visual}, which is a very large and the most commonly used dataset for the scene graph generation task. \cite{zhang2019graphical} focus on the ambiguous instances problem which includes the Entity Instance Confusion problem and the Proximal Relationship Ambiguity problem, and propose to tackle the above problems by utilizing the Graphical Contrastive Losses to explicitly force the model to disambiguate related and unrelated instances. \cite{zhan2019exploring,chen2019limitedlabels,peyre2019detectingunseen} choose to generate the labels based on external knowledge and internal existing labels to ease the problem of incomplete annotations.
Besides, \cite{zellers2018neuralmotif} analyzes and utilizes the motifs which are regularly appearing substructures in scene graphs to help detect relationships.
In the paper, we also follow the idea of exploring more information from existing labels. However, instead of directly analyzing the motifs
among labels, we build the predicate label hierarchy and further utilize the hierarchy to guide the feature learning.

\begin{figure*}[t]
\centering
\includegraphics[scale=0.36]{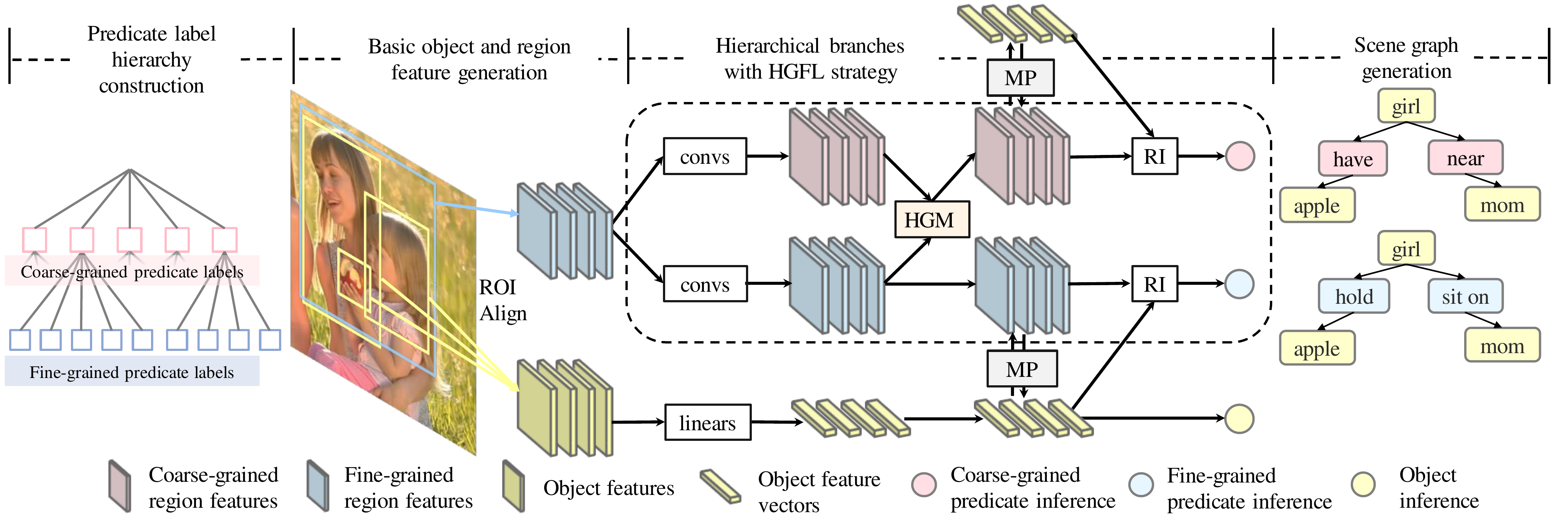}
\caption{\textbf{Overview of the Method.} The whole framework can be divided into four stages. 1) Predicate label hierarchy construction. 2) Basic object and region feature generation.
3) Hierarchical branches with HGFL strategy, in which HGM is settled to leverage the correlations among the coarse-grained and fine-grained region features. The Message Passing(MP) module is used to pass messages between object and region features. The Relation Inference(RI) module generates relationship features by fusing corresponding region and object features.
4) Scene graph generation. 
Best viewed in color.
}
\label{fig:method}
\end{figure*}

\subsubsection{Hierarchical knowledge.}  
Hierarchical information has been validated useful for many tasks
\cite{chao2019hierarchical_control,yang2019hierarchical_embedding,bugatti2019hicore,chen2019learning}. But there are only very few works utilizing hierarchical information on the relationship detection related task. 
\citet{bugatti2019hicore} utilizes the hierarchical relationship between image class, image superclass, and object bounding boxes to predict the global class of image such as bookstore and backery. Our work is different from their work in three folds. First, their goal, predicting the global image class, is completely different from ours that generates a scene graph which requires prediction of both objects and their pair-wise relationships in the image.
Second, the basic hierarchical knowledge is different, they mainly focus on the image class and object bounding box levels while we build the predicate label hierarchy. Third, we leverage the correlation between region features of different granularities, which is not utilized in
their work.

\citet{yin2018zoom} also try to resolve ambiguous and noisy object and predicate annotations by building the Intra-Hierarchical trees (IH-tree). 
However, our approach is different from theirs in the way of clustering labels and using hierarchical information.
(1) Clustering approach: in \cite{yin2018zoom}, labels are clustered based on the same verb, preposition or adjective \textbf{regardless of their semantic meanings}, which may deteriorate the problem of semantic ambiguity. Our method
cluster the predicate labels based on the semantic meanings.
For example,  "on a", "on side of", "on end of"  will be all assigned to a parent class "on" in \cite{yin2018zoom} but 
clustered to different parent classes by ours. "on side of", "next to", and "by" have different parent classes in \cite{yin2018zoom}, but have the same parent class by ours. Our approach is better at handling semantic ambiguity. 
(2) Using hierarchical information: Yin et al.  only use the losses of different 
granularities
for the same feature, which does not distinguish features from different granularities.
In comparison, our HGFL learns deep features for each granularity independently so that they are distinguishable, and then our HGM refines features by utilizing the correlations among these distinguished features.

\subsubsection{Message passing module.}
In scene graph generation task, relationship is highly dependent on both object and region features as it is represented as a subject-predicate-object phrase triplet, which is generated based on object and region features. 
Therefore, message passing modules for object-object, object-region, region-region are extensively studied
\cite{xu2017scene,qi2019attentive_relational_network}. 
\cite{vaswani2017attention_is_all,kipf2016gcn} are two common and classic templates of attention modules.
However, all these message passing modules are performed on the same granularity level. Our work is complementary to these works. Specifically, we propose the
HGM
to pass messages between features on different granularities levels, so that the features on the coarse-grained level can provide abstract guidance for the features on the fine-grained level. And the features on the fine-grained level can offer more detailed information for the features on the coarse-grained level in return.

\section{Method}
\subsection{The Entire Framework}
\label{sec:framework}
An overview of our method shown in Figure \ref{fig:method} could be summarized as:
1) Construct the predicate label hierarchy.
2) Given an image with potential objects and predicates, Faster RCNN \cite{faster_rcnn} is first applied to extract the basic object and region features. 
The region feature, extracted for predicting the categories of predicates, refers to the feature of a certain region containing multiple objects.
3) 
The hierarchical branches, trained with HGFL strategy, contain two branches.
One branch extracts the region features named \textit{coarse-grained region features} for coarse-grained predicate prediction,
while another branch extracts the region features named \textit{fine-grained region features} for the fine-grained predicate prediction.
Between the two branches, our HGM is settled to leverage the correlations among the coarse-grained and fine-grained region features. Besides, following Fnet\cite{li2018factorizable}, object and region features pass messages through the Message Passing (MP) module and then the Relation Inference (RI) module generates relationship features by fusing the corresponding region and object features.
To guarantee the two branches extracting the required features, we force them to be supervised by their corresponding coarse-grained and fine-grained predicate labels respectively.
4) The coarse-grained and fine-grained predicate categories are predicted using corresponding \emph{region} features, and the object categories are predicted using the \emph{object} features. Then predicates and objects are formulated into the final scene graph. 

\subsection{Predicate label hierarchy construction}
The overall hierarchy is built in a bottom-up manner by clustering the correlated predicate labels into each independent parent class.
The parent classes serve as the coarse-grained predicate labels and the cleaned predicate labels serve as the fine-grained predicate labels. These two hierarchical sets of labels are utilized for designing losses for the coarse-grained and fine-grained region branches.
In this paper, we introduce two ways of building the predicate label hierarchy(human understanding and machine understanding). 
More details are in Section Hierarchy Construction.

\subsection{Hierarchy Guided Feature Learning(HGFL)}
\label{sec:strategy}
Based on the built hierarchy, 
where the coarse-grained labels contain the semantic-overlapping connections between predicates and the fine-grained labels have the detailed information of predicates, we design the hierarchical branches to learn better region features of corresponding granularities. 
In the hierarchical branches,
given the features extracted from the ROI Align process, 
$2$ branches (coarse-grained and fine-grained region branch)
are used for 
extracting the corresponding coarse-grained region features 
 $\mathbf{A}$ 
and fine-grained region features 
 $\mathbf{B}$.
Then HGM is performed between $\mathbf{A}$ and $\mathbf{B}$ to exploit the correlations and utilize it to refine $\mathbf{B}$.
After HGM, MP module is used between each kind of region features and object features for refining features of each other. Then RI module generates relationship features by fusing region features and corresponding object features.

For our HGFL strategy, one cross-entropy loss is calculated in the fine-grained region branch using the fine-grained predicate labels, and another cross-entropy loss is applied in the coarse-grained region branch using the coarse-grained predicate labels.
In this way, the features of different branches correspond to predicates of different levels.
Note that when the network is implemented without the HGM, the coarse-grained branch can be removed during inference. This indicates that the improvement of our strategy is actually a freebie because our model will not introduce any extra computational expenses for deployment.

\begin{figure}
\centering
\includegraphics[scale=0.56]{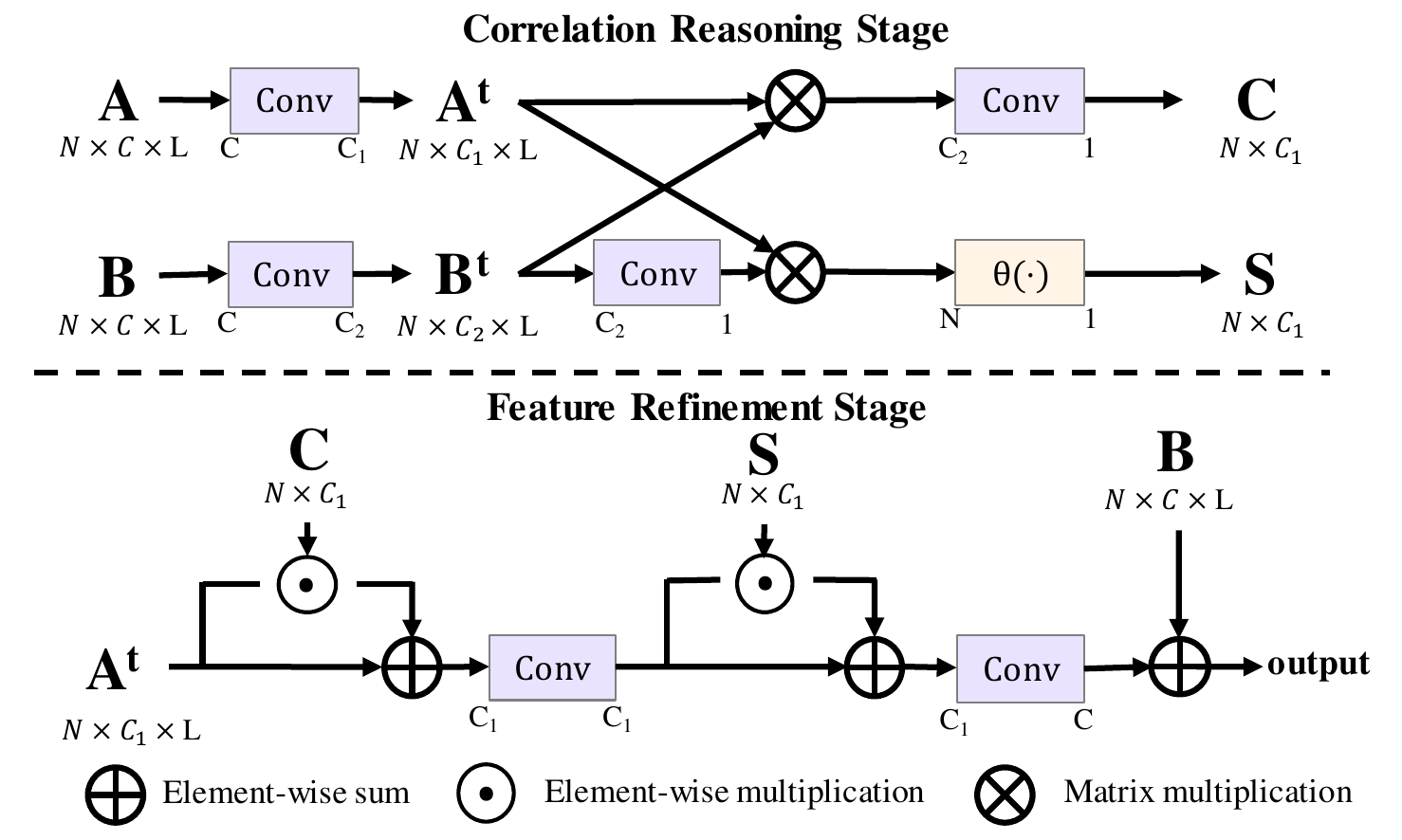}
\caption{\textbf{Structure of HGM.} 
\textbf{Top}: the Correlation Reasoning Stage, which reasons both region-wise and channel-wise correlations.  \textbf{Bottom}: the Feature Refinement Stage, which back-projects the reasoned correlations to the original feature space and utilize the back-projected correlations to refine the region features. Note that the notation $\textrm{Conv}$ in the figure actually represents the (Conv-BN-ReLU) triplet.}
\label{fig:hierarchy_guided_module}
\end{figure}

\subsection{Hierarchy Guided Module (HGM)}
\label{sec:HGM}
Hierarchy Guided Module (HGM) is designed to leverage the correlations of the coarse-grained and fine-grained region features.
As shown in Figure \ref{fig:hierarchy_guided_module}, it contains the Correlation Reasoning Stage and the Feature Refinement Stage.
\subsubsection{Correlation Reasoning.}
In this stage, feature transformation is firstly applied before the region-wise and channel-wise correlation reasoning.

\textbf{Feature transformation.}
Denote the region features of the coarse-grained and fine-grained level by $\mathbf{A} \in \mathbb{R}_{+}^{N\times C\times L}$ and $\mathbf{B} \in \mathbb{R}_{+}^{N\times C\times L}$ respectively, where $N, C$, and $L$ respectively denote
the numbers of regions, channels, and pixels. 
Two separate $1 \times 1$ convolutional layers are performed to transform
$\mathbf{A}$ and $\mathbf{B}$ 
into the same embedding space. 

The whole transformation process can be formulated as: 
\begin{equation}
\begin{split}
    \mathbf{A}^{t} &= \sigma(f^t_a(\mathbf{A}, \mathbf{W}^t_{a})),  \mathbf{A}^t \in \mathbb{R}_{+}^{N\times C_1 \times L} \\
    \mathbf{B}^{t} &= \sigma(f^t_b(\mathbf{B}, \mathbf{W}^t_{b})), \mathbf{B}^t \in \mathbb{R}_{+}^{N\times C_2 \times L}, \\
\end{split}
\label{eq:Tranform}
\end{equation}
where the notation $\sigma(\cdot)$ represents the non-linear activation function like ReLU,
$f^t_a(\cdot; W^t_a)$ and $f^t_b(\cdot; W^t_b)$ respectively denote the convolution operations for transforming region features $\mathbf{A}$ and $\mathbf{B}$ as with learnable weights $\mathbf{W}^t_a$ and $\mathbf{W}^t_b$. 
$\mathbf{A}^t $ and $\mathbf{B}^t$ in Eq.~(\ref{eq:Tranform}) denote the corresponding transformed region features. The constant $C_1$ and $C_2$ are the numbers of channels of the transformed region features.

\textbf{Region-wise correlation reasoning.} 
The region-wise correlation reasoning infers the correlations between the coarse-grained  features of a region and the fine-grained features of all other regions in the same image. 
For example, the fine-grained features from regions with similar predicates (e.g. ``close to'' and ``around'') will have higher correlations with the coarse-grained features of a region (e.g. "near").

To reduce the computational cost,
a convolution is applied to squeeze $\mathbf{B}^t\in \mathbb{R}_{+}^{N \times C_2 \times L}$ into just one channel as follows:
\begin{equation}
    \mathbf{B}_{sqz} = \sigma(f_{s}(\mathbf{B}^t; \mathbf{W}_{s})), \\
\end{equation}
where $f_{s}(\cdot; \mathbf{W}_{s})$ denote the convolution function with learnable weights $\mathbf{W}_{s}$, $\mathbf{B}_{sqz} \in \mathbb{R}_{+}^{N \times 1 \times L}$ denotes the output.  

Denote $\mathbf{A}^{t}_{i}\in \mathbb{R}_{+}^{C_1 \times L}$ as the $i$th region in $\mathbf{A}^t\in \mathbb{R}_{+}^{N \times C_1 \times L}$, where $i=1,\ldots,N$. 
To calculate the region-wise correlations, denoted as $\mathbf{S}$, between all channels of each region feature in $\mathbf{A}^t$ and the accumulated  $\mathbf{B}_{sqz}$, 
the $\mathbf{B}_{sqz}$ will firstly be reshaped into matrix $\mathbf{B}_{sqz}^M \in \mathbb{R}_{+}^{N \times L}$.

Then the following formulation is used:
\begin{equation}
\begin{split}
    \mathbf{S}_i = \theta(\mathbf{A}^{t}_{i} \times (\mathbf{B}_{sqz}^M)^{T}) \textrm{ for}\ i = 1, ..., N,  \\
\end{split}
\label{eq:S_i}
\end{equation}
where $\times$ denotes matrix multiplication, $(\cdot)^T$ denotes the transpose operation, and the function $\theta(\cdot)$ denotes the down-sampling function implemented by max or average pooling. 

In Eq.~(\ref{eq:S_i}), there are two steps. \underline{First}, matrix multiplication $\tilde{\mathbf{S}}_i=\mathbf{A}^{t}_{i} \times (\mathbf{B}_{sqz}^M)^{T}$ is performed for each region $\mathbf{A}^{t}_{i}$ to obtain the full region-wise correlation $\tilde{\mathbf{S}}_i$ between $\mathbf{A}^{t}_{i}$ and $\mathbf{B}_{sqz}^M$, and the size of $\tilde{\mathbf{S}}_i$ is $(C_1 \times N)$.
\underline{Second}, to further get the most related region of $\mathbf{B}^t$ for each region of $\mathbf{A}^t$, down-sampling operation $\theta(\cdot)$ is applied to reduce the second dimension of $\tilde{\mathbf{S}}_i$ from $N$ to $1$. Therefore, the size of $\mathbf{S}_i$ is $(C_1)$. 

The output $\mathbf{S}_i$ ($i= 1,\ldots,N$) corresponds to the correlation between the $i$th region in the coarse-grained features and all the other regions in the  fine-grained features. 
When all regions in the coarse-grained features are considered, the size of the full region-wise correlations  $\mathbf{S}$
is $(N \times C_1)$.

\textbf{Channel-wise correlation reasoning.}
In the region-wise correlation reasoning above, all channels of $\mathbf{B}^t$ are squeezed before calculating the correlation, which may lead to loss of channel information. 
To preserve and utilize the channel information, 
and to focus on the correlations between the coarse-grained and fine-grained features of the same region, 
we also calculate the channel-wise correlations.

Denote $\mathbf{B}^{t}_j\in \mathbb{R}_{+}^{C_2\times L}$ and $\mathbf{A}^{t}_j\in \mathbb{R}_{+}^{C_1\times L}$ as the $j$th region in $\mathbf{B}^{t}$ and $\mathbf{A}^{t}$ respectively. The channel-wise correlation matrix $\mathbf{C}_j$  of the $j$th region is obtained as follows:
\begin{equation}
\begin{split}
    &\mathbf{C}_j = \sigma(f_{c}(\mathbf{B}^{t}_j \times (\mathbf{A}^t_j)^T; \mathbf{W}_{c})) \textrm{ for}\ j = 1, ..., N, \\
\end{split}
\label{eq:C_j}
\end{equation}
where $\sigma(\cdot)$ represents the activation function and $f_{c}(\cdot;\mathbf{W}_{c})$ denotes a 1D-convolutional function with learneable weights $\mathbf{W}_{c}$.
There are three steps in Eq.~(\ref{eq:C_j}). \underline{First},
the correlation matrix $\tilde{\mathbf{C}}_j \in \mathbb{R}_{+}^{C_2 \times C_1}$ for each corresponding $j$th region in $\mathbf{A}^t$ and $\mathbf{B}^t$ can be obtained through matrix multiplication $\mathbf{B}^{t}_j \times (\mathbf{A}^t_j)^T$. 
\underline{Second}, 
to further globally gather and map the information in channels of each region in $\mathbf{B}^t$ to the domain of $\mathbf{A}^t$, the 1D-convolutional function $f_{c}(\cdot;\mathbf{W}_{c})$ is applied to reduce the first dimension of $\tilde{\mathbf{C}}_j$ from $C_2$ to 1. In this way, the size of $\mathbf{C}_j$ is $(1\times C_1)$. 
\underline{Third}, activation function $\sigma(\cdot)$ is applied.
Note that $\mathbf{C}_j$ is for the $j$th region, thus the size of whole channel-wise correlation $\mathbf{C}=[\mathbf{C}_1\ \ldots \mathbf{C}_j \ldots \ \mathbf{C}_{N}]$ 
is $(N \times C_1)$ considering all regions $j= 1,\ldots, N$.

\subsubsection{Feature refinement.}
In the feature refinement stage, the calculated region-wise and channel-wise correlations are firstly back projected into the region feature space and then stacked residual structures are utilized for feature refinement.
There are three steps in this stage. 

\underline{First},
denote $\mathbf{A}^{t}_l\in \mathbb{R}_{+}^{N \times C_1}$ as the $l$th pixel in $\mathbf{A}^{t}$,
the channel-wise correlations $\mathbf{C}$ is projected into the space of $\mathbf{A}^t_l$ through element-wise multiplication $\mathbf{A}^t_l * \mathbf{C}$, then the projected information is utilized as:
\begin{equation}
\label{eq:aout}
\begin{split}
    \mathbf{A}^{out}_l= \sigma(f_{ac}(\mathbf{A}^t_l + (\mathbf{A}^t_l * \mathbf{C});\mathbf{W}_{ac})), &\textrm{for}\ l=1,...,L, \\
\end{split}
\end{equation}
where $*$ denotes the element-wise multiplication, the function $f_{ac}(\cdot;\mathbf{W}_{ac})$ denotes the convolutional operation with learnable weights $\mathbf{W}_{ac}$.
Element-wise sum is utilized to fuse the $\mathbf{A}^t_l$ with the projected information $\mathbf{A}^t_l * \mathbf{C}$.
Then the convolution operation $f_{ac}(\cdot;\mathbf{W}_{ac})$ further non-linearly processes the fused features to the feature ${\mathbf{A}}^{out}_l$. 
The size of $\mathbf{A}^{out}= [\mathbf{A}^{out}_1, ..., \mathbf{A}^{out}_L]$ 
is ${(N \times C_1 \times L)}$ for all pixels $l= 1,\ldots, L$.

\underline{Second}, similar to Eq.~(\ref{eq:aout}), the region-wise correlations $\mathbf{S}$ is utilized
as follows:
\begin{equation}
\begin{split}
    \mathbf{A}^{out'}_l &= \sigma(f_{bs}(\mathbf{A}^{out}_l + (\mathbf{A}^{out}_l * \mathbf{S}); \mathbf{W}_{bs})), \textrm{for}\ l=1, ..., L, \\
\end{split}
\label{eq:aout2}
\end{equation}
where $f_{bs}(\cdot;\mathbf{W}_{bs})$ denote the convolutional function with learnable weights $\mathbf{W}_{bs}$.
The size of $\mathbf{A}^{out'}=[\mathbf{A}^{out'}_1\ldots \mathbf{A}^{out'}_l\ldots \mathbf{A}^{out'}_L]$ 
is $(N\times C \times L)$ for all pixels $l= 1,\ldots, L$.
The role of $f_{bs}(\cdot;\mathbf{W}_{bs})$ is to transform the fused features 
for refining the fine-grained region feature $\mathbf{B}$.

\underline{Third}, the $\mathbf{A}^{out'}$ serves as the correlated information from coarse-grained region features
and is added to the fine-grained region features $\mathbf{B}$ as follows: 
\begin{equation}
\begin{split}
    \mathbf{B}^{out} &= \mathbf{B} + \mathbf{A}^{out'}, \\
\end{split}
\end{equation}
where the refined feature $\mathbf{B}^{out}$ refers to the output of HGM.

\section{Hierarchy Construction}
\label{sec:dataset}
We introduce two clustering methods to build the predicate label hierarchy, 
Manual Clustering (MC) based on human understanding and Automatic Clustering (AC) based on machine understanding.
The clustered parent classes serve as the coarse-grained level and the cleaned predicate labels serve as the fine-grained level in the hierarchy.

\subsection{Manual Clustering}
In this process, clusters are made based on
the understanding of the text meaning of labels, scenes in the corresponding image and the synsets in Wordnet \cite{miller1995wordnet}.

For phrase-format predicate labels which include multiple words, we extract and consider the keyword based on the criterion whether the word can reflect the relation between subject and object or not.
Specifically, these 
labels can be categorised into three types, including verb-prep. (\emph{e.g.} "walking by"), prep.-prep. (\emph{e.g.} "in between") and stereotyped expression (\emph{e.g.} "inside of", "out of").
For the verb-prep., the verb will be chosen as the keyword only if it results in a static status, for example, we pick "topped" and "covered" as the keyword of "topped with" and "covered by".
Otherwise, the verb usually depicts the action and can be taken as the attribute of subject, so the prep. which usually contains the relative spatial relations will be taken as the keyword such as "in", "at" are the key role of "riding in", "sitting at" respectively.
For the prep.-prep. and stereotyped expressions, the corresponding image scene examples are taken as the important evidence for deciding the keyword because it is hard to tell the keyword from the text directly. 
\begin{table}
\setlength{\tabcolsep}{4.66pt}
\centering
\scalebox{0.66}{
\begin{tabular}{c|c|c|cc|cc|c|c}
\hline
\multirow{2}{*}{\makecell{Dataset}} &
\multirow{2}{*}{\makecell{\#Img}} &
\multirow{2}{*}{\makecell{\#Rel}} &
\multicolumn{2}{c|}{\textbf{Training Set}}  &
\multicolumn{2}{c|}{\textbf{Testing set}} & 
\multirow{2}{*}{\makecell{\#Obj}} &
\multirow{2}{*}{\makecell{\#Pred}} \\
& & & \#Img & \#Rel & \#Img & \#Rel & &\\
  \hline
  \makecell{VG-H} & 96235 & 771495 & 67364 & 540999 &   28871 & 230496 & 150 & 275(30) \\
  \hline
  \makecell{VG-MSDN} & 56164 & 618692 & 46164 & 507296 & 10000 & 111396 & 150 & 50 \\
  \hline
  \makecell{VG-DR-Net} & 76081 & 825405 & 67086 & 798906 & 8995 & 26499 & 399 & 24 \\ 
  \hline
\end{tabular}}
\caption{\textbf{Comparison among 
the cleaned VG-H, VG-MSDN and VG-DR-Net dataset.} '\#Img' and '\#Rel' refers to the number of image and relationships.
'\#Obj' and '\#Pred' represents the number of object and predicate categories respectively. Note that '\#Pred' for VG-H is shown in the format of "the number of fine-grained labels
(the number of coarse-grained labels)
".
}
\label{table:dataset}
\end{table}
\subsection{Automatic Clustering}
This method clusters the predicate labels based on the machine understanding through utilizing the pretrained word2vec\cite{word2vec} model.
Specifically, 
each predicate label is encoded as one $300$-dimensional embedding vector with the word2vec pretrained model.
Then Kmeans is applied to cluster these embedding vectors into parent classes. 
Because this method clusters features by comparing the calculated distances between embedding features, xivthe advantage is that the verb-prep. style relationship labels with the same preposition (\emph{e.g.} “sit on”, “walk on”, “stand on”) are easily clustered into same parent class because they have the same word. 
While the disadvantage is that 
the model will mainly focus on the pure text information or language prior instead of 
corresponding scene information.
For example, "park", "park near" and "near" are clustered into one parent class
because both "park" and "near" are closely related to "park near", 
but "park" and "near" are actually not semantically close.

\section{Experiments}
\subsection{Implementation details}
\textbf{Dataset.}
The Visual Genome (VG) dataset is selected as the base dataset due to its great capacity (108K images) and complexity (35 objects and 21 pairwise relationships within each image).
Before the hierarchy construction, we firstly perform the fundamental cleaning operations including filtering low-frequency predicate labels, removing or correcting meaningless predicate labels (\emph{e.g.} "no") and merging replaceable predicate labels (\emph{e.g.} merge "alongside" and "are alongside" into "alongside"). By applying the above cleaning method, the cleaned predicate labeling set has 275 categories. The clustered parent classes set has 30 categories.

We evaluate our results based on the cleaned predicate labels described above
if not specified. This dataset is denoted as VG-H. 
Except for VG-H, we also evaluate our method on other datasets, \textit{e.g.} VG-MSDN, VG-DR-Net.  
Comparison among these datasets is shown in Table 
\ref{table:dataset}. 

\noindent\textbf{Technical details of the framework.} 
The whole network is trained end-to-end,
VGG16 \cite{vgg} is selected as the backbone to extract basic features. 
As for the normalization within the whole framework, we apply synchronized BatchNorm only onto the HGM and the object bounding box head. During training, the number of object and region proposals are set to be 256 and 512. The number of channels of all object and region features are set to be 512. 
Both weights of two cross-entropy losses on coarse-grained and fine-grained region branches are set to 1.
\begin{table}
\centering
\scalebox{0.7}{
\begin{tabular}{c|c|cc|cc|cc} 
\hline
\multirow{2}{*}{\makecell{Dataset}} &
\multirow{2}{*}{\makecell{Name}} & 
\multicolumn{2}{c|}{\textbf{PredDet}}  &
\multicolumn{2}{c|}{\textbf{PhrDet}} & 
\multicolumn{2}{c}{\textbf{SGGen}}  \\
 & & $R^{50}$ & $R^{100}$ & $R^{50}$ & $R^{100}$ & $R^{50}$ & $R^{100}$ \\
 \hline
  \multirow{3}{*}{\makecell{Ours (VG-H)}} & baseline & 15.24 & 19.80 & 16.67 & 20.97 & 8.59 & 10.92 \\  
  & + HGFL & \textbf{18.06} & \textbf{22.84} & \textbf{21.26} & \textbf{26.42} & \textbf{11.36} & \textbf{14.14} \\  
  & + HGFL\_HGM & \textbf{19.44} & \textbf{24.01} & \textbf{21.63} & \textbf{26.36}  & \textbf{11.95} & \textbf{14.47} \\ 
 \hline
  \multirow{3}{*}{VG-MSDN} & baseline & 17.71 & 23.02 & 20.56 & 25.52 & 11.07 & 13.91 \\  
  & + HGFL & \textbf{20.10} & \textbf{26.35} & \textbf{24.39} & \textbf{30.40} &\textbf{13.17} & \textbf{16.90} \\  
  & +  HGFL\_HGM & \textbf{22.69} & \textbf{28.65} & \textbf{25.54} & \textbf{31.32}  & \textbf{14.82} & \textbf{18.34}  \\ 
  \hline
  \multirow{3}{*}{\makecell{VG-DR-Net}} & baseline & 29.35 & 35.16 & 27.06 & 33.23 & 18.32 & 22.07  \\ 
  & + HGFL & \textbf{37.58} & \textbf{44.51} & \textbf{33.51} & \textbf{40.04} & \textbf{24.57} & \textbf{29.22} \\  
  & + HGFL\_HGM & \textbf{39.12} & \textbf{46.98} & \textbf{33.67} & \textbf{40.66}  & \textbf{25.56} & \textbf{30.93}  \\ 
  \hline
\end{tabular} }
\caption{\textbf{Experiment results of component analysis on the cleaned dataset (VG-H), VG-MSDN and VG-DR-Net.}
\emph{PredDet}, {\emph{PhrDet}}, {\emph{SGGen}}
denote Predicate Detection, {Phrase Detection}, and {Scene Graph Detection}
,$R^{K}$ denotes the Recall for the top $K$ predictions, which also applies for all other Tables in this paper.}
\label{table:component_analysis}
\end{table}

\noindent\textbf{Evaluation.}
The model is evaluated on three tasks, 
all assuming \textit{ground-truth object bounding boxes are not provided}, 
Predicate Detection (PredDet), Visual Phrase Detection (PhrDet) and Scene Graph Generation (SGGen). Predicate Detection aims to detect the predicate category based on the RPN proposal. 
Visual Phrase Detection is to detect the subject-predicate-object phrases requiring the IoU (intersection over union) value between region bounding box and ground truth region bounding box should be at least 0.5.
Scene Graph Generation is also to detect both the objects and their pairwise relationships, but it requires the IoU between object pairs and the ground truth to be higher than 0.5.

The Top-K Recall(Recall@K) is chosen as the main evaluation metric. It calculates the fraction of ground-truth relationships hit in the top K predictions. K is set to 50 and 100 in our evaluation. 

\subsection{Component analysis}
For component analysis, we perform experiments on the proposed
HGFL strategy and the HGM
on the datasets including the proposed VG-H dataset, VG-MSDN and VG-DR-Net. 
Fnet \cite{li2018factorizable} is taken as the base framework. 

\noindent \textbf{Hierarchy Guided Feature Learning (HGFL).}
Experiment results of 
HGFL strategy
on the VG-H dataset, VG-MSDN and VG-DR-Net are shown in Table \ref{table:component_analysis}.
The baseline shown in the first row refers to 
training the Fnet \cite{lin2017fpn} only with the fine-grained predicate labels, and the result "+HGFL" refers to training the model with both the coarse-grained and fine-grained predicate labels through our HGFL strategy. 
As we can see, on VG-H dataset,  the results of Recall@50 and Recall@100 can respectively achieve 18.5\% and 15.35\% gain on PredDet, 27.53\% and 25.98\% gain on PhrDet, 32.24\% and 29.84\% gain on SGGen. 
Similar gains are achieved in the experiment results on the VG-MSDN and VG-DR-Net datasets.
All above results show that HGFL strategy is a simple but very effective method.

\subsubsection{Hierarchy Guided Module (HGM).}
HGM is used for reasoning the cross-correlations between coarse-grained and fine-grained region features, which is complementary to HGFL strategy. 
Since other methods do not have coarse-grained predictions, to fairly compare with them,
the output of HGM is the refined fine-grained region features.
As shown in Table \ref{table:component_analysis}, 
"+HGFL\_HGM" denotes adding HGM and training the model with HGFL strategy. Compared with the model trained only with HGFL strategy, on VG-H dataset, the pure gain of HGM on the Recall@50 and Recall@100 can achieve 7.64\% and 5.12\% on PredDet, 5.19\% and 2.33\% on SGGen.
Besides, on VG-H dataset, the overall gain of both
HGFL and HGM on Recall@50 and Recall@100 can achieve 27.55\% and 21.26\% on PredDet, 29.75\% and 25.7\% on PhrDet, 39.11\% and 32.5\% on SGGen. 
\begin{table}
 \centering
 \scalebox{0.8}{
\begin{tabular}{c|cc|cc|cc}
\hline
\makecell{Name} & 
\multicolumn{2}{c|}{\textbf{PredDet}}  &
\multicolumn{2}{c|}{\textbf{PhrDet}} & 
\multicolumn{2}{c}{\textbf{SGGen}} \\
(Groups) & $R^{50}$ & $R^{100}$ & $R^{50}$ & $R^{100}$ & $R^{50}$ & $R^{100}$ \\  
  \hline
  MC(30) & 18.06 & 22.84 & 21.26 & \textbf{26.42} & 11.36 & 14.14  \\ 
  \hline
  AC(30) & \textbf{18.22} & \textbf{22.91} & \textbf{21.30} & 26.22 & \textbf{11.59} & \textbf{14.22}  \\
  \hline
  AC(50) & 18.12 & 22.84 & 21.10 & 26.03 & 11.52 & 14.24  \\
  \hline
  AC(100) & 17.97 & 22.85 & 21.04 & 25.91 & 11.32 & 14.05  \\
  \hline
\end{tabular} }
\caption{\textbf{Comparison between 
Manual Clustering (MC) and Automatic Clustering (AC) with different number of groups (clusters)
on VG-H dataset.} 
MC and AC
have comparable performance, which validates that hierarchy 
construction
is consistently useful for refining features.
} 
\label{table:clustering_method}
\end{table}

\subsection{Ablation study}
\label{sec:ablation_study}
\begin{table}
\setlength{\tabcolsep}{5.66pt}
\centering
\scalebox{0.8}{
\begin{tabular}{c|cc|cc|cc} 
\hline
\makecell{Name} & 
\multicolumn{2}{c|}{\textbf{PredDet}}  &
\multicolumn{2}{c|}{\textbf{PhrDet}} & 
\multicolumn{2}{c}{\textbf{SGGen}} \\ 
& $R^{50}$ & $R^{100}$ & $R^{50}$ & $R^{100}$ & $R^{50}$ & $R^{100}$ \\
\hline
 \makecell{HGFL} & 18.06 & 22.84 & 21.26 & \textbf{26.42} & 11.36 & 14.14  \\  
  \hline
  \makecell{HGFL\_concat} & 17.43 & 22.03 & 20.57 & 25.22 & 11.07 & 13.68  \\  
  \hline
  \makecell{HGFL\_MPS} & 18.06 & 22.78 & 21.06 & 26.085 & 11.45 & 14.10  \\ 
  \hline
  \makecell{HGFL\_HGM} & \textbf{19.44} & \textbf{24.01} & \textbf{21.63} & 26.36  & \textbf{11.95} & \textbf{14.47}  \\ 
  \hline
\end{tabular}}
\caption{\textbf{Comparison of different message passing modules between coarse-grained and fine-grained branches.} Experiment results show that HGM has better performance over other general message passing methods. 
}
\label{table:message_passing_module}
\end{table}
\noindent \textbf{Hierarchy construction method.}
Experiment results on Manual Clustering (MC) and Automatic Clustering (AC) are shown in the first two rows of Table \ref{table:clustering_method}, 
the conclusion is that these two methods have comparable performance. Our analysis is as follows: the verb-preposition style relationship account for a large portion of all relationship labels, and the preposition is usually taken as the keyword when clustering, thus the phenomenon that most verb-preposition relationship labels with the same preposition (\emph{e.g.} "stand next to", "lying next to") will be clustered into the same group (\emph{e.g.} "next to") is 
the common property of these two methods. 
While the difference between these two methods is that the MC is mainly based on human understanding but the 
AC is mainly based on the
machine understanding which comes from the pretrained model's language prior.
However, it is very difficult for model to correctly fit the cluster related to human understanding or pure 
machine understanding, 
but it is easy for the model to capture the \emph{common regularity} of clustering most verb-preposition relationship labels with the same preposition into one group.
Besides, the comparable gains on both MC and AC also confirm that the method of hierarchy construction
and employment is effective.

\noindent \textbf{Clustering group number.}
When building the hierarchy manually, the number of clustering groups(30) is not pre-defined but the result of manually clustering.
Therefore, the number of clustering groups in AC is a factor to be explored. We perform experiments on 30, 50, and 100 respectively. Experiment results are shown in the last three rows of Table \ref{table:clustering_method}, 
which show that the number of clustering groups is not that important. And there is slight performance drop when the number of clusters increases. 
\begin{table}
\setlength{\tabcolsep}{6.66pt}
\scalebox{0.55}{
\begin{tabular}{c|c|cc|cc}
\hline
\makecell{Dataset} & 
\makecell{Model} & 
\multicolumn{2}{c|}{\textbf{PhrDet}} & 
\multicolumn{2}{c}{\textbf{SGGen}} \\
 & & $R^{50}$ & $R^{100}$ & $R^{50}$ & $R^{100}$ \\
\hline
 \multirow{3}{*}{\makecell{VG-DR-Net \\
 \cite{dai2017detecting}}}
 & DR-Net \cite{dai2017detecting} & 23.95 & 27.57 & 20.79 & 23.76 \\ 
 & Fnet:2-SMP \cite{li2018factorizable} & 26.91& 32.63 & 19.88& 23.95 \\ 
 & Ours  & \textbf{33.67}& \textbf{40.66} & \textbf{25.57}& \textbf{30.94} \\
 \hline
 \multirow{6}{*}{\makecell{VG-MSDN \\ \cite{li2017MSDN}}} 
 & \makecell{ISGG\cite{xu2017scene}} & 15.87& 19.45 & 8.23& 10.88 \\
 & MSDN \cite{li2017MSDN} & 19.95& 24.93 & 10.72& 14.22 \\
 & Fnet: 2-SMP \cite{li2018factorizable} & 22.84& 28.57 & 13.06& 16.47 \\
 & KB-GAN \cite{gu2019scene} & 23.51& 30.04 & 13.65& 17.57 \\
 & Ours & \textbf{25.55}& \textbf{31.33} & \textbf{14.83}& \textbf{18.34} \\
  \hline
\end{tabular}}
\caption{\textbf{Comparison with state-of-the-art methods on different datasets.} The results show that the proposed method outperform the state-of-the-art methods. 
}
\label{table:different_dataset}
\end{table}

\noindent \textbf{Message passing module.} The experiment results are shown in Table \ref{table:message_passing_module}. The baseline Fnet model plus our HGFL (HGFL in Table \ref{table:message_passing_module}) is used as the baseline in this experiment. The investigated message passing methods include \cite{vaswani2017attention_is_all} (HGFL\_MPS in Table \ref{table:message_passing_module}), directly concatenating both region features in the coarse-grained and fine-grained level (HGFL\_concat in Table \ref{table:message_passing_module}), and HGM (HGFL\_HGM in Table \ref{table:message_passing_module}). All these message passing methods are applied on the same position of model while training with the HGFL strategy. As shown in Table \ref{table:message_passing_module}, the performance of general message passing method or the concatenation does not provide improvement compared with the baseline HGFL. And HGM provides improvement on all evaluation metrics, which shows that HGM is effective.

\subsection{Comparison with the state-of-the-art.}
Our method could be generalized for any framework and any dataset, in which labels have the inherent semantic-overlapping connections. 

\noindent \textbf{Different datasets.} Since the results based on hierarchy from the MC and AC are similar, we directly build hierarchy through AC for other datasets which include VG-MSDN \cite{li2017MSDN} and VG-DR-Net \cite{dai2017vg-drnet}. The clustering group number is set to 8 and 6 respectively. The experiment results are shown in Table \ref{table:different_dataset}. As we can see in the table, our method outperforms the state-of-the-art approaches on both datasets. Results on VG-MSDN and VG-DR-Net also show that our baseline Fnet is a strong baseline approach.
Besides, compared with the IH-Tree \cite{yin2018zoom},
experimental results show that our method is more effective. For VG dataset, our pure averaged gain of Recall(@50 and @100) among three tasks is 4.27\%, while the corresponding gain of IH-Tree \cite{yin2018zoom} is 2.01\%.

\begin{table}
\setlength{\tabcolsep}{1.66pt}
\centering
\scalebox{0.66}{
\begin{tabular}{c|c|cc|cc|cc} 
\hline
\multirow{2}{*}{\makecell{Framework}} &
\multirow{2}{*}{\makecell{Name}} & 
\multicolumn{2}{c|}{\textbf{PredDet}}  &
\multicolumn{2}{c|}{\textbf{PhrDet}} & 
\multicolumn{2}{c}{\textbf{SGGen}} \\
 & & $R^{50}$ & $R^{100}$ & $R^{50}$ & $R^{100}$ & $R^{50}$ & $R^{100}$ \\
 \hline
  \multirow{2}{*}{ISGG \cite{xu2017scene}} & baseline & 13.96 & 17.67 & 16.72 & 20.86 & 8.42 & 10.44 \\ 
  & +  ours & \textbf{15.54} & \textbf{19.62} & \textbf{18.16} & \textbf{22.19}  & \textbf{9.35} & \textbf{11.45} \\ 
 \hline
  \multirow{2}{*}{MSDN \cite{li2017MSDN}} & baseline & 13.24 & 16.63 & 16.15 & 20.22 & 7.81 & 9.62 \\ 
  & + ours & \textbf{14.79} & \textbf{18.95} & \textbf{18.20} & \textbf{22.53}  &
  \textbf{9.24} & \textbf{11.51} \\ 
  \hline
  \multirow{2}{*}{\makecell{Graph-RCNN \cite{yang2018graph}}} & baseline & 13.29 & 16.57 & 15.77 & 19.56 & 7.52 & 9.23 \\ 
  & + ours & \textbf{17.42} & \textbf{22.27} & \textbf{18.18} & \textbf{23.02} & \textbf{10.09} & \textbf{12.76} \\  
  \hline
  \multirow{2}{*}{\makecell{Fnet \cite{li2018factorizable}}} & baseline & 15.24 & 19.80 & 16.67 & 20.97 & 8.59 & 10.92 \\ 
  & + ours & \textbf{19.44} & \textbf{24.01} & \textbf{21.63} & \textbf{26.36}  & \textbf{11.95} & \textbf{14.47} \\ 
  \hline
\end{tabular}
}
\\
\caption{\textbf{Experimental results on the different frameworks on the cleaned dataset (VG-H).}
Our method can further improve the performance on different frameworks.
}
\label{table:component_analysis_mf}
\end{table}

\noindent \textbf{Different frameworks.} In Table \ref{table:component_analysis_mf}, we further provide the results of applying our method onto different frameworks which include ISGG \cite{xu2017scene}, MSDN \cite{li2017MSDN} and Graph-RCNN \cite{yang2018graph} on the VG-H dataset. The Fnet \cite{li2018factorizable} is the original baseline in this paper. 
All these frameworks are reimplemented. In the table, for each framework, 
"+ ours" denotes the results obtained by training the baseline with our method (the HGFL strategy and the HGM). Based on the experimental results, we find that our method can further improve the performance on different frameworks.  

\section{Conclusions}
In this paper, 
to explore the semantic-overlapping connections in the predicate labels, 
we firstly propose to build the language hierarchy, which contains both the coarse-grained and fine-grained levels, based on comprehensive semantic understanding in two different perspectives including human understanding and machine understanding.
Then we introduce a Hierarchy Guided Feature Learning strategy to learn better region features of the two levels. Besides, we further propose the Hierarchy Guided Module to better utilize the cross-level correlations. Experiment results show that our method is a general method, which not only outperforms the baseline on the predicate labeling sets but also outperforms state-of-the-art methods on other public datasets.



%
%
\bibliography{egbib}
\end{document}